# Classification of worldwide news articles by perceived quality, 2018-2024

Connor McElroy
Department of Computer Science
Lakehead University
Orillia, Canada
csmcelro@lakeheadu.ca

Thiago E. A. de Oliveira
Department of Computer Science
Lakehead University
Orillia, Canada
talvesd@lakeheadu.ca

Chris Brogly
Department of Information Technology
Algoma University
Brampton, Canada
christopher.brogly@algomau.ca

*Abstract*— This study explored whether supervised machine learning and deep learning models can effectively distinguish perceived lower-quality news articles from perceived higher-quality news articles. 3 machine learning classifiers and 3 deep learning models were assessed using a newly created dataset of 1,412,272 English news articles from the Common Crawl over 2018-2024. Expert consensus ratings on 579 source websites were split at the median, creating perceived low and high-quality classes of about 706,000 articles each, with 194 linguistic features per website-level labelled article. Traditional machine learning classifiers such as the Random Forest demonstrated capable performance (0.7355 accuracy, 0.8131 ROC AUC). For deep learning, ModernBERT-large (256 context length) achieved the best performance (0.8744 accuracy; 0.9593 ROC-AUC; 0.8739 F1), followed by DistilBERT-base (512 context length) at 0.8685 accuracy and 0.9554 ROC-AUC. DistilBERT-base (256 context length) reached 0.8478 accuracy and 0.9407 ROC-AUC, while ModernBERT-base (256 context length) attained 0.8569 accuracy and 0.9470 ROC-AUC. These results suggest that the perceived quality of worldwide news articles can be effectively differentiated by traditional CPU-based machine learning classifiers and deep learning classifiers.

## I. Introduction

Given the vast amount of news sources on the web and the daily volume of news articles published, there is a need to be able to recognize the perceived quality of this content at a large scale. Continuous human evaluation of article quality has become impractical, creating a need for automated perceived quality detection methods. In this paper, we examine whether we can differentiate perceived lower-quality news articles from perceived higher-quality news articles using machine learning (ML) classifiers and deep learning. Using an overall quality score gathered from expert evaluations of news websites in prior research [1], 1,412,272 articles were matched with website-level labels as either low or high quality for binary classification. 194 NLP features were derived from article text to evaluate if article style could aid in differentiating quality via machine learning classifiers. Three deep learning models were also evaluated: DistilBERT [2] (with context lengths of 256 and 512 tokens), ModernBERT-base, and ModernBERT-large [2] (context length of 256 tokens). The evaluation of perceived quality here focused on looking at NLP features and deep learning predictions of website-level rather than article-level labelled examples from a trusted source, or, in other words, a perceived quality; our approach does not directly evaluate content for factuality or reliability, which tends to be the focus of fake news classification.

Differentiating text by perceived quality has important implications. There are many prior works on classification of fake news [3][4][5][6] but not necessarily on article quality. For researchers, automated quality detection based on NLP features or deep learning predictions is a convenient way to analyze and filter datasets of news articles. Additionally, these models could serve as a tool that may be of interest to readers about the perceived quality of the content that they browse.

## II. Related Work

### A. Perceived Text Quality

Work on perceived text quality has moved past surface readability toward linguistic aspects that align with human judgment. Prior work has shown that models using lexical, syntactic, and discourse features track human judgment better than length-based formulas, showing that discourse structure and coherence are key predictors of perceived quality [7]. Subsequent studies in journalistic writing demonstrate that article-level quality is learnable from textual and stylistic cues [8]. Within the news domain, fluency and completeness have been identified as the strongest indicators of perceived quality [9].

Recent work has explored the use of human-machine detector output as unsupervised indicators of language quality [10]. In this [10] methodology, "page quality" and "language quality" are used interchangeably and detector scores are validated against human-rated perceived quality categories. The approach uses detector outputs as text-based proxies for low language quality. In contrast, our study utilizes a supervised framework that derives perceived article quality from domain-level expert consensus, using these scores as estimated labels for article-level models. This approach captures the writing patterns characteristic of news websites. A previous dataset in news, published yearly, also provided website-level ratings [11], although in [11] classification was not done directly on the data. In that work, content was manually scraped in comparison to being taken from Common Crawl News as it is here.

## B. Website-Level Quality Assessment

Automated assessment of perceived article quality requires scalable solutions beyond manual review. Two approaches to assessment include article-level quality labelling and website-level quality labelling. Article-level is preferable, although it is difficult to rate tens of thousands of news articles for quality manually. On the other hand, outlets such as Media Bias Fact Check and AllFact, among others, can provide assessments of the quality of various news domains [1], which could be used for website-level ratings. However, because there are different organizations evaluating domains using different scoring systems and criteria, a decision has to be made to select one of them, although a single score taking into account all of these would be beneficial.

Our work is enabled by the findings published by Lin et al. [1], who compared six different sets of website-level expert ratings and found a high level of similarity among them. Because the expert sources generally agreed on which domains were higher or lower quality, this work used imputation and Principal Component Analysis (PCA) to generate aggregate quality ratings, called PC1 scores.

These PC1 scores measure website-level quality ranging from 0.0 (lowest) to 1.0 (highest), capturing signals common to different sources related to factual accuracy, bias assessment, and review scores. The resulting PC1 ratings are an aggregate, or "wisdom-of-experts" [1] approach for a single measure of quality [1].

## C. Summary and Research Gap

Domain-level quality ratings such as the PC1 scores established by Lin et al. [1] provide an ideal aggregated measurement of perceived news quality, although they do not capture the linguistic and stylistic variability of article content within domains. Existing work has excelled at measuring quality across outlets but has not explored whether the same signal can be predicted directly from article text. This gap limits the scalability of quality assessment on the level of article content.

Our study builds on these efforts by using the PC1 scores as estimated labels to train classifiers at scale. Through large-scale supervised learning, we test whether linguistic features and transformers can recover website-level perceived quality from article text.

## III. METHODS

### A. Introduction

This section describes the methods used for data collection and model training. First, a web-content parser isolated relevant article text using structural and text heuristics. Next, dataset collection, language filtering, NLP feature extraction, and assignment of standardized quality scores were performed. Traditional supervised machine learning classifiers were then trained and validated to distinguish content quality based on 194 NLP features. Finally, transformer-based deep learning models (DistilBERT and ModernBERT) were fine-tuned to evaluate performance for binary perceived quality classification.

### B. Parsing Implementation

The parser design is built with a combination of text heuristics and structural indicators to obtain relevant article text content. For instance, it removes cookie consent nodes, unrelated HTML tags, and script blocks, then scans each candidate section <div>, <section>, <article>, and <main>. Each candidate is scored based on paragraph count, word count, link density, and class or ID attributes that could indicate article-like content. Paragraphs contribute +2 points each, and overall text length adds +1 point per 10 words, assuming that longer, paragraph-rich sections are more likely to represent article bodies. Sections with excessive links receive a –10 penalty, while those whose class or ID attributes include terms such as article, content, body, or story gain a +100 boost. Conversely, sections tagged as comment, footer, sidebar, or scrollbar are penalized by –500, discouraging selection of irrelevant areas. Typical article sections achieve total scores in the 150–300 range, while navigational or footer sections usually remain below 50. Within the highest scoring section, raw text and paragraph nodes are refined through rules to skip copyright disclaimers and short text segments. By scoring nodes based on these parameters, the parser locates the main body of the article with a high degree of confidence.

### C. Data Cleaning

For this study, we built an entirely new dataset of online news articles. The dataset consists of 1,412,272 rows of article text with source website from the Common Crawl News database over 2018-2024. For each article text, we also include 194 NLP features gathered using the SpaCy [12] library. The previously discussed related study on aggregate news website-level quality was needed for data labelling, so their provided ratings were used. These PC1 scores (the first principal component of a PCA on an imputed dataset on news website-level quality [1]) were assigned as binary labels for every article in our dataset, each of which is from a website that matched one of the 579 website domains with a PC1 rating. PC1 scores are a measure of quality ranging from 0.0 (lowest quality) to 1.0 (highest quality) created from the consensus of several expert sources regarding overall website-level news quality. Binary classification using the median PC1 score of 0.8301 was applied as the threshold.

This same approach to labelling via the median was also used in another work from our group on classifying news headlines on the web [13]. Using multiple classes (> 3) resulted in less training samples for each class, so we opted for a binary classification setup to give each class a more representative sample size. We found that the median was a good separator of low quality to high quality text because PC1 scores such as 0.86 are considered as "relatively high" ratings in the initial paper [1].

### D. Supervised Machine Learning

CPU-based supervised learning was performed on 3 traditional classifiers. Other classifiers were initially included but removed due to slow train time and/or worse performance than Random Forest. All 194 NLP features (overviewed in Table 2) gathered for each article were used in the training process. 5-fold stratified cross validation was used to measure the consistency of each model's performance measures. A fixed seed was used to generate the folds and this was reused for all three classifiers.

TABLE I: MACHINE LEARNING MODELS & PARAMETERS

| Model | Parameters |
|---|---|
| Random Forest | n_estimators=200, class_weight="balanced", n_jobs=-1, random_state=42 |
| Logistic Regression | max_iter=1000, class_weight="balanced", random_state=42 |
| Gaussian Naïve Bayes | Default |
| DistilBERT (base-uncased) | batch_size = 32, grad_acc = 4, epochs = 4, learning_rate = 2 × $10^{-5}$, weight_decay = 0.01, optimizer = AdamW, scheduler = cosine decay (warmup_ratio = 0.1), max_seq_len = 512, fp16 = enabled (CUDA), 5-fold stratified CV (seed = 42). (discriminative LR: backbone = 1 × $10^{-5}$, head = 2 × $10^{-5}$) |
| ModernBERT-base | batch_size = 64, grad_acc = 2, epochs = 4, learning_rate = 2 × $10^{-5}$, weight_decay = 0.01, optimizer = AdamW, scheduler = cosine decay (warmup_ratio = 0.1), max_seq_len = 256, fp16 = enabled (CUDA), 5-fold stratified CV (seed = 42). (discriminative LR: backbone = 1 × $10^{-5}$, head = 2 × $10^{-5}$; FlashAttention enabled) |
| ModernBERT-large | batch_size = 32, grad_acc = 4, epochs = 4, learning_rate = 2 × $10^{-5}$, weight_decay = 0.01, optimizer = AdamW, scheduler = cosine decay (warmup_ratio = 0.1), max_seq_len = 256, fp16 = enabled (CUDA), 5-fold stratified CV (seed = 42). (discriminative LR: backbone = 1 × $10^{-5}$, head = 2 × $10^{-5}$; FlashAttention enabled) |

### E. Deep Learning Models

Three Deep Learning Models were developed using BERT variants: DistilBERT-base-uncased, ModernBERT-base, and ModernBERT-large. Experiments were run on Ubuntu 20.04.6 LTS with an AMD Ryzen 7 5700G 8-core processor, 121 GiB of RAM, and an NVIDIA GeForce RTX 4090 GPU with 24 GB of VRAM. All three were trained under a common PyTorch/Hugging Face pipeline including preprocessing, data splitting, optimization, scheduling, evaluation, logging, and model selection. Tokenization was done with each model's native tokenizer using truncation and padding to a fixed context length. The ModernBERT models were run at 256 tokens due to memory constraints, while DistilBERT was run at 256 and 512 tokens. Stratified 5-fold cross-validation was used with a fixed random seed, which enabled us to reuse the same folds for every model. In each fold, every model trained for four epochs with validation at the end of each epoch. Accuracy, ROC-AUC, F1, precision, and recall was computed and a checkpoint was saved every epoch. After training, the checkpoint with the minimum validation loss was restored and evaluated once on the fold's validation set to obtain the fold's metrics. Final cross-validation results report the mean ± standard deviation of these metrics.

The fine-tuning process was identical across each model. Optimization used AdamW with discriminative learning rates (classification head 2×$10^{-5}$, backbone 1×$10^{-5}$), weight decay of 0.01 on non-bias/LayerNorm parameters, a cosine schedule with 10% warmup, mixed precision (FP16), and gradual unfreezing (head trained first, all layers unfrozen after the first epoch). We fixed the effective batch size at 128 across each model run, with adjustments to the batch size and gradient accumulation to address out-of-memory (OOM) errors.

TABLE II: DATASET PROPERTIES

| Dataset Property | Value | Description |
|---|---|---|
| Initial articles | 10,904,097 | Full dataset size from Common Crawl source |
| English-only articles | 7,491,131 | Subset of articles detected as English |
| Total articles used | 1,412,272 | Rows matched with PC1-scored domains, final count |
| Unique news domains analyzed | 579 | The amount of PC1-scored domains |
| Median PC1 score value | 0.8301 | Selected for even class distribution |
| Number of part of speech (POS) features | 20 | Counts or ratios of POS tags (nouns, verbs, adjectives, etc.). |
| Number of Penn treebank features | 57 | Phrase-structure tag features derived from Penn treebank annotations (e.g., NP, VP) |
| Number of dependency label features | 72 | Features capturing syntactic dependency roles (e.g., nsubj, dobj) |
| Number of NER features | 26 | Frequency features for named-entity recognition categories (PERSON, ORG, DATE, etc.) |
| Number of additional NLP measures | 21 | Miscellaneous linguistic metrics |
| Total number of features per text | 194 | Aggregate count of all engineered features used as model input. |

## IV. Results

Basic dataset information is provided in Table 2. Table 3 reports five-fold cross-validation means with standard deviations. Figure 1 shows average loss per epoch for the top 3 models. The machine learning classifiers serve as baselines that allow us to assess the added value of deep learning models. Among these baselines, Random Forest performs the best (Accuracy 0.7355 ± 0.0013; ROC-AUC 0.8131 ± 0.0014; F1 0.7396 ± 0.0014). Deep learning models improve results substantially, as our top model overall is ModernBERT-large (Accuracy 0.8744 ± 0.0007; ROC-AUC 0.9593 ± 0.0004; F1 score 0.8739 ± 0.0006). Compared to Random Forest, ModernBERT-large improves accuracy by 13.89%, ROC-AUC by 14.62%, and F1 score by 13.43%. DistilBERT-base (512 tokens) also shows large gains over Random Forest (accuracy +13.30%, ROC-AUC +14.23%, F1 score +12.83%). Gains in the top-performing deep learning models come from two main factors: increasing DistilBERT's context from 256 to 512 tokens provides consistent improvements across all metrics (accuracy +2.07%, ROC-AUC +1.47%, F1 score +2.08%), and scaling up from ModernBERT-base to ModernBERT-large provides further performance gains (accuracy +1.75%, ROC-AUC +1.23%, F1 score +1.77%). These rankings are consistent with model capabilities and show that increasing context length is beneficial to overall model performance. Gaussian Naïve Bayes and Logistic Regression underperformed here, possibly because they didn't handle the high number of NLP features well in comparison to the Random Forest. The Random Forest performed better likely by capturing non-linear splits and interaction effects. The transformer-based models tested provided the largest gains in ROC-AUC and F1 Score. While the improvements might appear modest (10%+), that does translate to thousands of additional articles correctly classified. The leading models show balanced performance across evaluation metrics. ModernBERT-large achieves macro-averaged precision 0.8780 ± 0.0017 and recall 0.8732 ± 0.0005, indicating consistent accuracy across both high- and low-quality classes. The close alignment of precision and recall shows that the model maintains an even trade-off between false positives and false negatives rather than favoring one label. This balance, combined with very low variability across folds (≤ 0.0018), demonstrates stable generalization and reliable classification behavior across data splits.

FIGURE I: AVERAGE LOSS PER EPOCH FOR TOP 3 MODELS

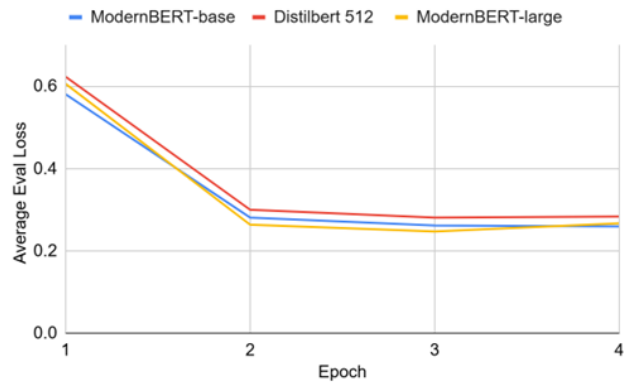

TABLE III: FIVE-FOLD CV PERFORMANCE (MEAN ± STD)

| Classifier/Model | Accuracy | ROC AUC | F1 Score | Precision | Recall |
|---|---|---|---|---|---|
| Gaussian NB | 0.5551 ± 0.0130 | 0.5911 ± 0.0008 | 0.4873 ± 0.1080 | 0.5971 ± 0.0265 | 0.4363 ± 0.1560 |
| Logistic Regression | 0.6359 ± 0.0013 | 0.6880 ± 0.0013 | 0.6461 ± 0.0014 | 0.6438 ± 0.0012 | 0.6483 ± 0.0019 |
| Random Forest | 0.7355 ± 0.0013 | 0.8131 ± 0.0014 | 0.7396 ± 0.0014 | 0.7464 ± 0.0012 | 0.7329 ± 0.0021 |
| DistilBERT-base-uncased 256 token | 0.8478 ± 0.0010 | 0.9407 ± 0.0005 | 0.8471 ± 0.0011 | 0.8509 ± 0.0009 | 0.8466 ± 0.0011 |
| DistilBERT-base-uncased 512 token | 0.8685 ± 0.0007 | 0.9554 ± 0.0004 | 0.8679 ± 0.0007 | 0.8711 ± 0.0009 | 0.8674 ± 0.0007 |
| ModernBERT-base | 0.8569 ± 0.0017 | 0.9470 ± 0.0011 | 0.8562 ± 0.0018 | 0.8601 ± 0.0015 | 0.8557 ± 0.0019 |
| ModernBERT-large | 0.8744 ± 0.0007 | **0.9593 ± 0.0004** | **0.8739 ± 0.0006** | 0.8780 ± 0.0017 | 0.8732 ± 0.0005 |

## V. Discussion

Each domain's overall quality PC1 score rating was applied to its respective articles, providing a broadly representative website-level measure of quality. Given that content on news sites often follow writing guidelines, it is reasonable to assume that the PC1 scores generally can reflect individual article quality as well, although there still may be variation in individual article quality vs. overall domain quality.

We openly admit that the website-level ratings applied to each article may not necessarily be perfect, but that article-level labelling is currently impractical with human evaluators. As a result, we argue this is a reasonable trade-off to still involve reliable expert evaluation. Our goal with this work was to match these website-level labels with a large number of NLP measures for machine learning and a large number of article-texts for deep learning. The improved performance of DistilBERT at a higher context length (512 tokens compared to 256 tokens) may suggest that capturing the complete sequence length of the article content enhances the model's discriminatory capability. Since DistilBERT's context length reaches its maximum at 512, we are unable to improve its performance via context length. Therefore, further experiments exploring extended context lengths with models like ModernBERT-base and ModernBERT-large are necessary to fully understand additional potential via context length.

One limitation of this work was that article quality is determined based strictly on NLP features or deep learning predictions, which does not check the accuracy or reliability of the information contained in the articles. Rather, the focus is on determining if the articles appear to be in-line with previously labelled examples of a perceived low or high-quality news article.

## VI. Conclusion

This analysis demonstrates that machine learning and deep learning methods can effectively differentiate perceived lower-quality news articles from perceived higher-quality articles. Using a newly constructed dataset from Common Crawl News consisting of 1,412,272 articles labelled via overall news domain quality PC1 scores, traditional machine learning classifiers and fine-tuned deep learning models showed good predictive capability. Among the deep learning models evaluated, ModernBERT-large (256 tokens) achieved the highest overall results (Accuracy = $0.8744 \pm 0.0007$, ROC AUC = $0.9593 \pm 0.0004$, F1 = $0.8739 \pm 0.0006$), surpassing both DistilBERT and ModernBERT-base across all evaluation metrics. DistilBERT also showed a minor performance improvement when increasing its context length from 256 to 512, suggesting that capturing full article text enhances model performance. Further experimentation with extended sequence lengths in ModernBERT-base and ModernBERT-large models would be beneficial with more hardware resources. Future opportunities may include exploring multi-class quality assessments to provide more granular evaluations beyond binary classifications, and further investigation in assessing the performance of additional state-of-the-art deep learning models.

## References

[1] H. Lin et al., "High level of correspondence across different news domain quality rating sets," *PNAS Nexus*, vol. 2, no. 9, p. pgad286, Sep. 2023, doi: 10.1093/pnasnexus/pgad286.

[2] B. Warner et al., "Smarter, Better, Faster, Longer: A Modern Bidirectional Encoder for Fast, Memory Efficient, and Long Context Finetuning and Inference," Dec. 2024, [Online]. Available: http://arxiv.org/abs/2412.13663

[3] N. Capuano, G. Fenza, V. Loia, and F. D. Nota, "Content-Based Fake News Detection With Machine and Deep Learning: a Systematic Review," Apr. 14, 2023, *Elsevier B.V.* doi: 10.1016/j.neucom.2023.02.005.

[4] J. Y. Khan, Md. T. I. Khondaker, S. Afroz, G. Uddin, and A. Iqbal, "A benchmark study of machine learning models for online fake news detection," *Machine Learning with Applications*, vol. 4, p. 100032, Jun. 2021, doi: 10.1016/j.mlwa.2021.100032.

[5] F. Farhangian, R. M. O. Cruz, and G. D. C. Cavalcanti, "Fake news detection: Taxonomy and comparative study," *Information Fusion*, vol. 103, Mar. 2024, doi: 10.1016/j.inffus.2023.102140.

[6] M. Q. Alnabhan and P. Branco, "Fake News Detection Using Deep Learning: A Systematic Literature Review," *IEEE Access*, vol. 12, pp. 114435–114459, 2024, doi: 10.1109/ACCESS.2024.3435497.

[7] E. Pitler and A. Nenkova, "Revisiting Readability: A Unified Framework for Predicting Text Quality," 2008.

[8] A. Louis and A. Nenkova, "What Makes Writing Great? First Experiments on Article Quality Prediction in the Science Journalism Domain." [Online]. Available: http://www.cis.upenn.edu/

[9] I. Arapakis, F. Peleja, B. Berkant, and J. Magalhaes, "Linguistic Benchmarks of Online News Article Quality," in *Proceedings of the 54th Annual Meeting of the Association for Computational Linguistics (Volume 1: Long Papers)*, K. Erk and N. A. Smith, Eds., Berlin, Germany: Association for Computational Linguistics, Aug. 2016, pp. 1893–1902. doi: 10.18653/v1/P16-1178.

[10] D. Bahri, Y. Tay, C. Zheng, C. Brunk, D. Metzler, and A. Tomkins, "Generative Models are Unsupervised Predictors of Page Quality: A Colossal-Scale Study," in *Proceedings of the 14th ACM International Conference on Web Search and Data Mining*, in WSDM '21. New York, NY, USA:

Association for Computing Machinery, 2021, pp. 301–309. doi: 10.1145/3437963.3441809.

[11] M. Gruppi, B. D. Horne, and S. Adalı, "NELA-GT-2022: A Large Multi-Labelled News Dataset for The Study of Misinformation in News Articles", doi: 10.7910/DVN/AMCV2H.

[12] "spaCy GitHub." [Online]. Available: https://github.com/explosion/spaCy

[13] A. Mccutcheon, T. E. A. De Oliveira, A. Zheleznov, and C. Brogly, "Binary classification for perceived quality of headlines and links on worldwide news websites, 2018-2024."